\documentclass[10pt,twocolumn,letterpaper]{article}

\usepackage[pagenumbers]{cvpr} 

\usepackage{multirow}
\usepackage{diagbox} 

\definecolor{cvprblue}{rgb}{0.21,0.49,0.74}
\usepackage[pagebackref,breaklinks,colorlinks,allcolors=cvprblue]{hyperref}

\def\paperID{*****} 
\def\confName{CVPR}
\def\confYear{2026}

\title{SATMapTR: Satellite Image Enhanced Online HD Map Construction\thanks{Code and data will be made publicly available at \url{https://github.com/chivas1000/SATMapTR-Satellite-Image-Enhanced-Online-HD-Map-Construction/tree/main}}}

\author{Bingyuan Huang\\
City University of Hong Kong\\
{\tt\small bihuang-c@my.cityu.edu.hk}
\and
Guanyi Zhao\\
City University of Hong Kong\\
{\tt\small guanyzhao3-c@my.cityu.edu.hk}
\and
Qian Xu\\
City University of Hong Kong\\
{\tt\small qian.xu@cityu.edu.hk}
\and
Yang Lou\\
City University of Hong Kong\\
{\tt\small yanglou3-c@my.cityu.edu.hk}
\and
Yung-Hui Li\\
Hon Hai Research Institute \\
{\tt\small yunghui.li@foxconn.com}
\and
Jianping Wang\\
City University of Hong Kong\\
{\tt\small jianwang@cityu.edu.hk}
}

\begin{document}
\maketitle
\begin{abstract}
High-definition (HD) maps are evolving from pre-annotated to real-time construction to better support autonomous driving in diverse scenarios. However, this process is hindered by low-quality input data caused by onboard sensors’ limited capability and frequent occlusions, leading to incomplete, noisy, or missing data, and thus reduced mapping accuracy and robustness.
Recent efforts have introduced satellite images as auxiliary input, offering a stable, wide-area view to complement the limited ego perspective. However, satellite images in Bird’s Eye View are often degraded by shadows and occlusions from vegetation and buildings. Prior methods using basic feature extraction and fusion remain ineffective.
To address these challenges, we propose SATMapTR, a novel online map construction model that effectively fuses satellite image through two key components: (1) a gated feature refinement module that adaptively filters satellite image features by integrating high-level semantics with low-level structural cues to extract high signal-to-noise ratio map-relevant representations; and (2) a geometry-aware fusion module that consistently fuse satellite and BEV features at a grid-to-grid level, minimizing interference from irrelevant regions and low-quality inputs.
Experimental results on the nuScenes dataset show that SATMapTR achieves the highest mean average precision (mAP) of 73.8, outperforming state-of-the-art satellite-enhanced models by up to 14.2 mAP. It also shows lower mAP degradation under adverse weather and sensor failures, and achieves nearly 3$\times$ higher mAP at extended perception ranges.
\end{abstract}    
\section{Introduction}


\begin{figure}[t]
\centering
\includegraphics[width=1.0\linewidth]{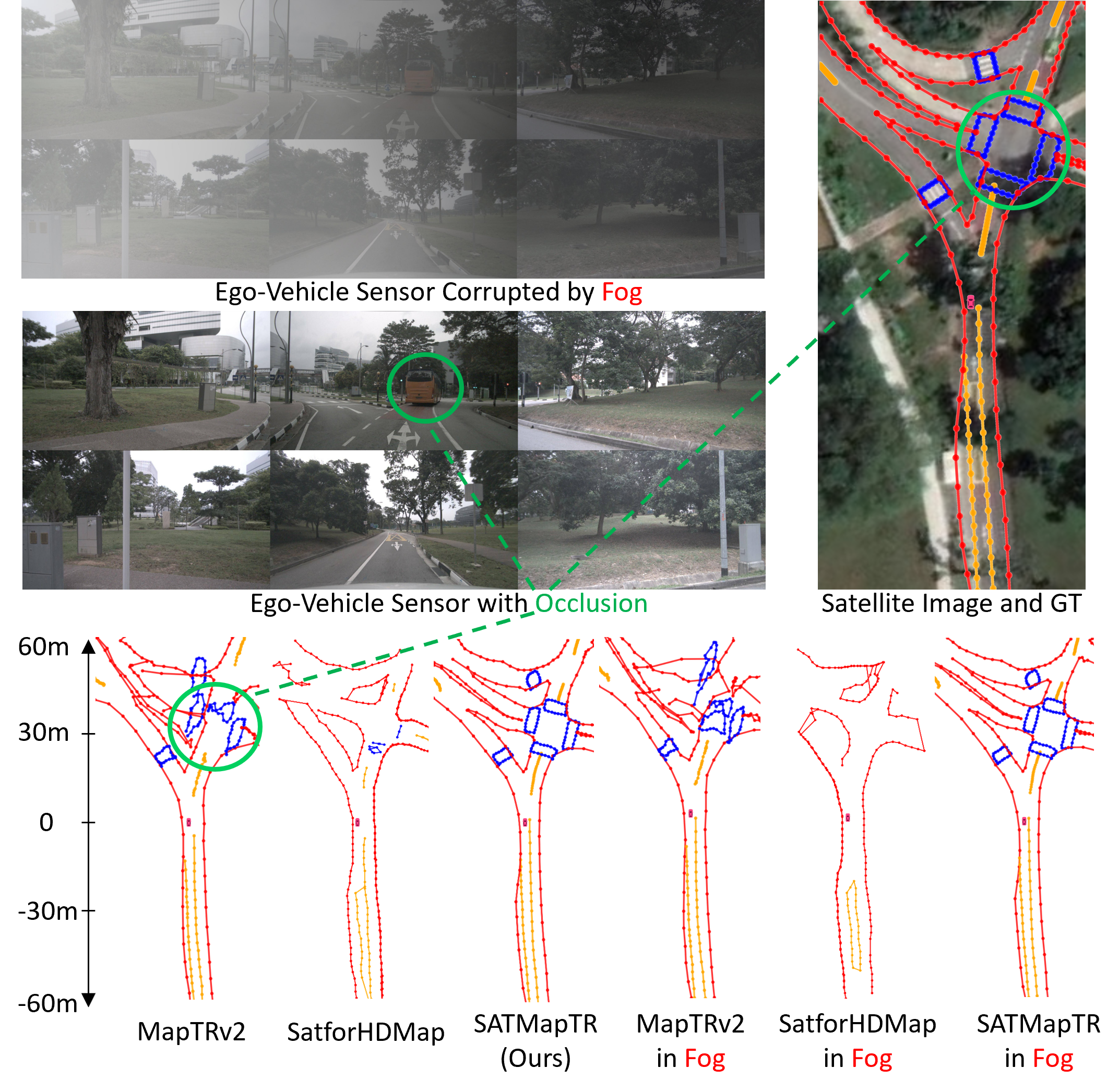}
\caption{Comparison of HD map construction results of different models under normal and challenging fog conditions in extended perception range ($-60\,\mathrm{m}$ to $60\,\mathrm{m}$). The green circle denotes regions of foreground object occlusion and the corresponding affected areas in HD map predictions.}
\label{fig:satforhdmap_challenging_comparison}
\end{figure}

\begin{figure*}[htbp!]
\centering
\includegraphics[width=\textwidth]{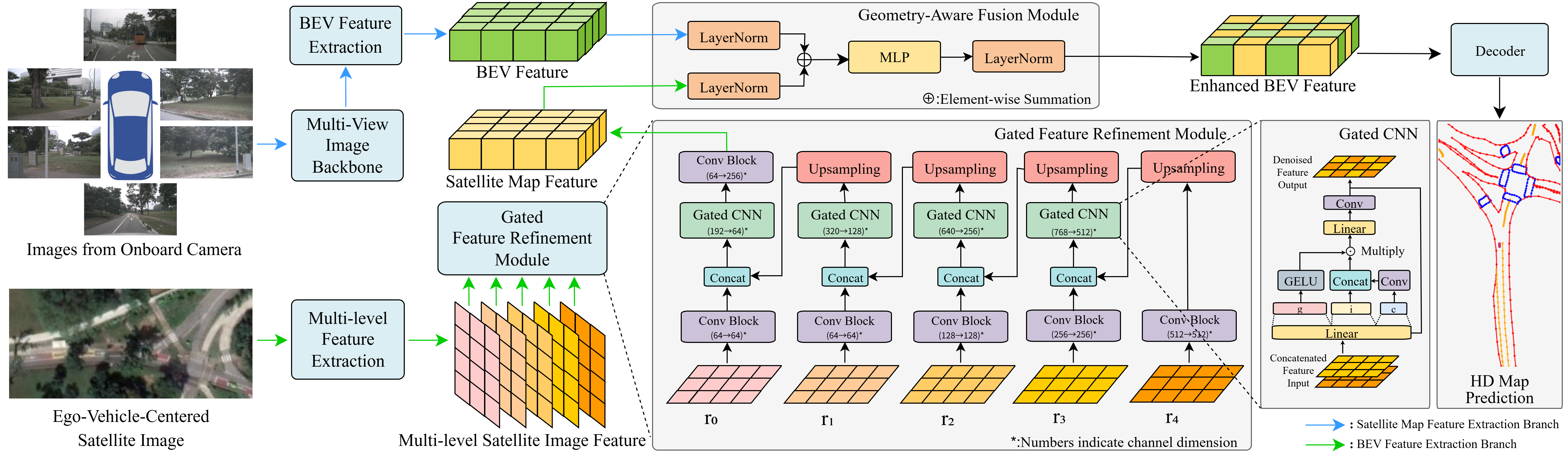}
\caption{Overall Architecture}
\label{fig:architecture}
\end{figure*}

High-definition (HD) maps are essential for autonomous driving, providing high-accuracy map elements such as road boundaries, lane dividers, and pedestrian crossings. Traditional HD maps are constructed offline through complex pipelines involving point cloud collection, SLAM-based map building, and manual annotation. Recently, online HD map construction approaches like~\cite{li2022hdmapnetonlinehdmap, liu2023vectormapnetendtoendvectorizedhd, liao2022maptr} utilize onboard sensor data to to obtain BEV features~\cite{li2023delvingdevilsbirdseyeviewperception} and construct vectorized HD maps for downstream autonomous driving tasks.


Online HD mapping approaches hold promise for real-time, automated updates, yet they face substantial challenges in terms of input data quality in achieving accuracy and robustness across various environments. These challenges are largely due to the inherent constraints of onboard sensor data. First, sensor limitations—such as limited sensing range, vulnerability to environmental changes, and possible failures. Second, occlusion from the ego-vehicle's perspective, where foreground objects block visibility, results in missed or incorrect detections of map elements. These sensor limitations and occlusions not only impact common scenarios but also significantly reduce the quality of input data in extensive perception ranges and challenging conditions. Therefore, utilizing auxiliary data sources alongside onboard perception is crucial to improving map accuracy and robustness.

Satellite images and Standard-definition (SD) maps, which are publicly available resources, have been widely used as auxiliary input to enhance online map construction. However, SD maps typically represent roads only as centerlines without boundary details or lane markings~\cite{jiang2024pmapnetfarseeingmapgenerator,wu2024blosbevnavigationmapenhanced}, limiting their effectiveness in improving HD map construction. 
In contrast, satellite images not only capture structural information but also provide rich semantic details (see supplementary materials~\ref{sec:sdsatcom}), such as road boundaries, lane markings, and pedestrian crossings, which more closely align with the requirements of HD maps for autonomous driving systems.

While previous research~\cite{gao2024complementingonboardsensorssatellite} has demonstrated the potential of satellite imagery to improve online HD map construction, substantial opportunities still exist for enhancing both accuracy and robustness across diverse scenarios. Through a comprehensive analysis of leveraging satellite image in online HD map construction, we identify the following key challenges:

\textbf{Challenge 1: Satellite Image Denoise \& Feature Extraction.} Clouds, shadows, vegetation, and buildings frequently occlude critical road elements in satellite imagery, introducing substantial noise. In this context, we define \textit{satellite map features} as features closely related to map elements—such as road boundaries, pedestrian crossings, and lane markings—distinguishing them from generic satellite image features that may contain irrelevant information. Conventional CNN-based methods~\cite{ronneberger2015unetconvolutionalnetworksbiomedical, lin2017featurepyramidnetworksobject} process all features uniformly, making it challenging to suppress noise or selectively enhance useful features. Consequently, developing methods for the efficient extraction of satellite map features from noisy satellite image remains a critical unsolved challenge.

\textbf{Challenge 2: Geometry-Aware Satellite-BEV Feature Fusion.} While satellite imagery and BEV features share the same top-down perspective, existing attention-based fusion approaches lack explicit spatial constraints and rely on input quality. BEV features are inherently location-specific, with each grid encoding information for a specific spatial region. Preserving this spatial correspondence during early-stage fusion is crucial to maintain the interpretability and effectiveness of the enhanced BEV features, ensuring that the representations remain aligned with subsequent decoding and processing. Attention-based fusion struggles to integrate complementary representation when either feature is degraded, as degraded feature cannot effectively match corresponding map elements and aggregate them from complementary features. This leads to unreliable fusion with a lack of meaningful complementarity (see supplementary materials~\ref{sec:limitation_ca}). Effective fusion should also remaining immune to degraded inputs.
Therefore, developing methods to accurately and consistently fuse satellite map features with BEV features across diverse condition while preserving the fidelity of location-specific information remains a critical challenge.

To address these challenges, we propose SATMapTR, a novel end-to-end framework that establishes a complete processing pipeline from raw surrounding view images, augmented by satellite imagery, to vectorized HD map elements for accurate and robust satellite-BEV feature extraction and fusion in online HD map construction. We propose a gated feature refinement module that aggregates multi-level satellite image features, refining from high-level semantics to low-level structural information, while selectively retaining map-relevant information. Our geometry-aware fusion module enforce explicit fusion with distance constraints, ensuring features are exclusively fused from the same physical locations, thereby eliminating interference from irrelevant or degraded regions, thus enhancing the quality of BEV representations. As shown in Fig.~\ref{fig:satforhdmap_challenging_comparison}, SATMapTR achieves superior mapping accuracy and robustness compared with previous methods, demonstrating effectiveness not only under normal situations, but also in the presence of occlusions, extended perception ranges—typically limited by the sensing range of onboard sensors—and in challenging environments, such as adverse weather and lighting. These results validate the effectiveness of our design for real-time HD map construction in practical autonomous driving applications. By leveraging satellite images, SATMapTR provides a comprehensive solution and sets a new standard for robust and accurate HD mapping.

The main contributions of this work are as follows: 
\begin{itemize}
\item We propose a gated feature refinement module that leverages multi-level feature aggregation and gated units to suppress irrelevant noise and selectively enhance map-relevant signals in satellite imagery.
\item We introduce a geometry-aware fusion module that applies explicit spatial distance constraints for grid-to-grid fusion, ensuring accurate and consistent feature fusion strictly between corresponding physical locations of satellite and BEV spatial coverage.
\item Extensive experiments on the nuScenes dataset demonstrate that SATMapTR outperforms state-of-the-art vision-based and satellite-enhanced approaches. Notably, it excels in large perception ranges and maintains robust performance under challenging conditions.
\end{itemize}

\section{Related Work}
\subsection{Online HD Map Construction} 
Solutions can be broadly categorized into detection-based~\cite{wang2018lanenetrealtimelanedetection, 5548087} and segmentation-based methods~\cite{8237796, neven2018endtoendlanedetectioninstance, chen2022efficientrobust2dtobevrepresentation, zhou2022crossviewtransformersrealtimemapview, li2022bevformerlearningbirdseyeviewrepresentation, liu2024bevfusionmultitaskmultisensorfusion}.
HDMapNet~\cite{li2022hdmapnetonlinehdmap} employed semantic segmentation on BEV feature to generate rasterized maps and turns it into vectorized map via post-processing. Subsequently, VectorMapNet~\cite{liu2023vectormapnetendtoendvectorizedhd} marked a breakthrough by utilizing Transformers to accomplish map element detection and polyline prediction tasks, making it first capable of predicting vectorized HD maps in end-to-end ways. Building upon this progress, MapTR~\cite{liao2022maptr} innovatively utilized Transformers for more ingenious instance and point-level matching tasks, further improving real-time efficiency and mapping performance through a query-based detection paradigm. Furthermore, its successor MapTRv2~\cite{liao2024maptrv2endtoendframeworkonline} further introduced auxiliary supervision, decoupled self-attention and one-to-many matching scheme, achieving superior detection accuracy. Recently, more research
such as PivotNet~\cite{ding2023pivotnetvectorizedpivotlearning} and MapQR~\cite{liu2024leveragingenhancedqueriespoint} has introduced novel solutions focusing on key point detection and query optimization respectively, further advancing the field.

Despite these advancements, existing methods still struggle with occlusion and long-distance perception. Moreover, system robustness is limited as it relies solely on onboard sensors, See also~\cite{li2023delvingdevilsbirdseyeviewperception}.

\subsection{Online HD Map Construction with Auxiliary Input}
Recent approaches increasingly leverage auxiliary inputs to enhance HD map construction. Some use self-constructed map priors~\cite{zhang2024enhancingvectorizedmapperception}, others utilize SD map data to improve BEV features~\cite{jiang2024pmapnetfarseeingmapgenerator,wu2024blosbevnavigationmapenhanced,luo2023augmentinglaneperceptiontopology}, while some pioneer the use of satellite imagery~\cite{gao2024complementingonboardsensorssatellite}. However, efficient feature extraction and fusion of auxiliary input remain open challenges. Our work advances this direction through adaptive extraction and precise fusion of satellite map information, which breaks through onboard sensor limitations and significantly improves the quality of online HD map construction.
\section{Method}
We propose a novel satellite image-enhanced framework for online HD map construction, designed to significantly improve mapping accuracy and robustness. As shown in Figure~\ref{fig:architecture}, our architecture consists of four core components: (1) BEV Feature Extraction Branch, (2) Satellite Map Feature Extraction Branch, (3) Geometry-Aware Fusion Module, and (4) Map Decoder. 
For BEV feature extraction branch, we adopt Lift-Splat-Shoot (LSS)~\cite{philion2020liftsplatshootencoding} to transform multi-view camera images into a unified BEV representation $F_{\text{bev}}$.
The satellite map feature extraction branch addresses Challenge 1, satellite image noise, by employing our novel gated multi-level architecture, specifically designed to (1) suppress noise from clouds and vegetation and (2) selectively enhance map-relevant features.
We then perform geometry-aware fusion to address Challenge 2, achieving accurate and consistent fusion with spatial constraints.
Finally, we implement the map decoder from~\cite{liao2024maptrv2endtoendframeworkonline} to process the fused features and generate the final HD map predictions.

\subsection{Satellite Map Feature Extraction Branch}
\label{sec:sat_feat_extract}
In this branch, we extract map-relevant satellite map features from the the processed satellite imagery using a hierarchical approach. The pipeline begins by aligning the satellite image with the BEV space. After alignment, satellite image features are extracted at multiple levels. Finally, a gated feature refinement module is applied to reduce map-irrelevant noise in the extracted satellite map features.

\subsubsection{Satellite Image Alignment.}
To obtain a satellite image spatially aligned with the BEV perception space, we first convert the ego-vehicle’s local pose into global geographic coordinates and orientation. Based on this global pose, We then crop a satellite image patch centered at the ego-vehicle from a large satellite image, covering the pre-defined BEV perception range (i.e., $60\,\mathrm{m} \times 30\,\mathrm{m}$ in physical dimensions), and resize it to a fixed resolution ($I_{sat} \in \mathbb{R}^{H_{sat} \times W_{sat} \times 3}$).
Next, we transform the satellite image patch from the global geographic coordinate system to the ego-vehicle coordinate system using an affine transformation. In the nuScenes dataset, for instance, this transformation consists of a vertical flip followed by a $90^\circ$ clockwise rotation, (see supplementary materials~\ref{sec:satalignment}). This ensures that the satellite image and BEV features are coarsely aligned in the same coordinate system. Beyond this initial alignment, the subsequent Gated Feature Refinement module further achieves fine-grained spatial alignment, which enables more accurate feature extraction and fusion.

\subsubsection{Multi-level Feature Extraction.} \label{sec:feat_extract}
Satellite imagery contains rich hierarchical information, ranging from high-level semantic patterns (e.g., lane types) to low-level structural details (e.g., lane contours and road signs). We extract multi-level image features $r_0, r_1, \dots, r_4$ from the satellite image, which serve as a foundation for constructing semantically coherent and structurally informative satellite map feature. The extraction process can be formulated as:
\begin{equation}
\begin{aligned}
  &r_0 = \text{Interp}\Big(\text{MaxPool}\big(\text{Conv}(I_{\text{sat}})\big)\Big) \\
  &\ (r_1, r_2, r_3, r_4) = \text{ResNet}(I_{\text{sat}})
\end{aligned}
\label{eq:sat_feat_extract}
\end{equation}

Here, $I_{\text{sat}}$  denotes the input satellite image, and $r_0, r_1, r_2, r_3, r_4$ represent multi-level image feature outputs from a ResNet-based backbone. 
$r_0$ is obtained by applying a convolution and max-pooling to $I_{\text{sat}}$, followed by bilinear upsampling to the BEV resolution $H \times W$, resulting in a feature preserves fine structural information. Features $r_1$ to $r_4$ are extracted from main stages of the ResNet backbone, capturing progressively higher-level semantic details. These multi-level image features ${\{r_0, r_1, r_2, r_3, r_4\}}$ are then refined by our Gated Feature Refinement module to obtain satellite map feature.

\begin{figure}[!t]
\centering
\includegraphics[width=1.0\linewidth]{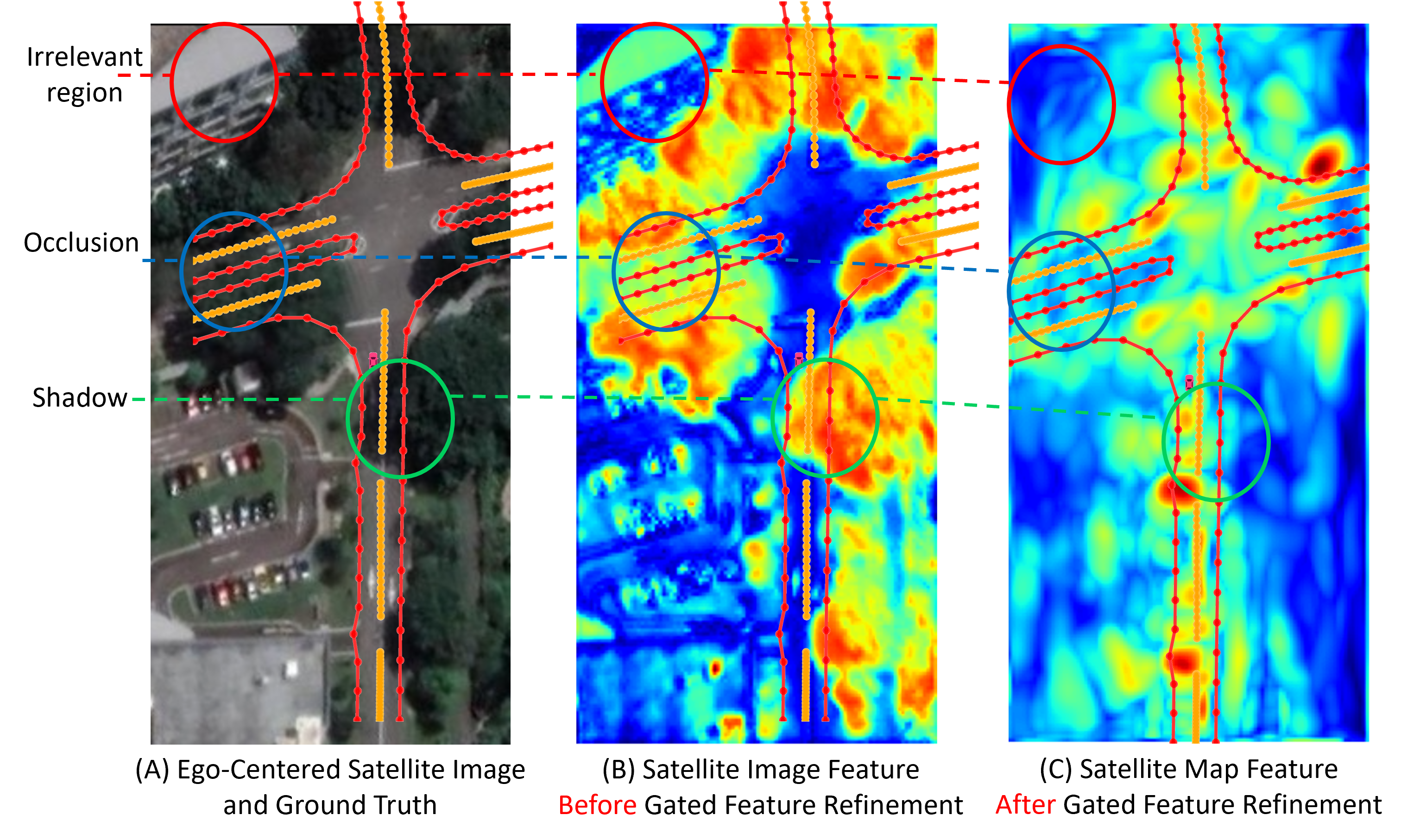}
\caption{Visualization of the Gated Feature Refinement module. Activation strength increases from blue (low) to red (high). Circle in red denotes region irrelevant to map elements; blue denotes area where map elements are occluded; green denotes map element covered by shadows. As shown in (B), the input satellite image features are not aligned with the GT HD Map, and the circled regions exhibit high responses due to noise. After GFR, the output satellite map features in (C) exhibit noise filtering and aligned responses with the GT, demonstrating GFR's misalignment compensation and noise suppression.}
\label{fig:gate_multi_scale_satellite_map_feature_refinement}
\end{figure}

\subsubsection{Gated Feature Refinement module.} \label{sec:feat_refinement}
This module addresses Challenge 1: the inherent noise and irrelevant information (e.g., shadows, occlusions, cloud cover) in raw satellite imagery degrade the map accuracy. We propose a Feature Refinement Module that implements two key operations: (1) Semantic-Structural Information Aggregation—integrating high-level semantic cues with low-level structural details to preserve both context and geometry; (2) Adaptive Feature Selection—dynamically filtering out unreliable or irrelevant noises while amplifying map-relevant signals through learned gating mechanisms. The module operates on the principle of content-aware filtering, where the gating weights are conditioned on both local feature confidence and global context, ensuring robust feature representation for downstream tasks.

As illustrated in Fig.~\ref{fig:architecture}, each satellite image feature at different level is first processed by a convolutional block, which aggregates local neighborhood information. Since the features are already coarsely aligned, this local aggregation enables the network to further refine the alignment at a finer scale. The convolution thus serves as an implicit fine alignment, allowing the network to adaptively compensate for small misalignments between the satellite and BEV grids, as shown in Fig.~\ref{fig:gate_multi_scale_satellite_map_feature_refinement}, the refined satellite map features exhibit improved alignment with the ground-truth HD map.

Building on this fine alignment, our refinement pipeline hierarchically processes features, starting from the deepest $r_4$. At each stage, the higher-level feature is upsampled, concatenated with the adjacent lower-level feature, and passed through a gated CNN for selective enhancement. This process iterates from $r_4$ to $r_0$, producing the refined satellite map feature.

\textbf{Upsampling:} The feature $r_i$ is upsampled via bi-linear interpolation, denoted as $r_i^{\text{in}}$, to match the resolution of feature $r_{i-1}$, outputs ${\hat{r}_i}$, the upsampling stage can be expressed as: 
\begin{equation}
\begin{aligned}
  &\hat{r}_i = \textit{Interp}(r_i^{\text{in}})
\end{aligned}
\label{eq:interpolation}
\end{equation}

\textbf{Concatenation:} The upsampled feature ${\hat{r}_i}$ is concatenated with the lower-level feature $r_{i-1}$. The output is denoted as $\tilde{r}_{i-1}$. 
\begin{equation}
\begin{aligned}
  \tilde{r}_{i-1} = \mathrm{Concat}(\hat{r}_i, r_{i-1})
\end{aligned}
\label{eq:Concatenation}
\end{equation}

\textbf{Gated CNN Processing:} To mitigate noise propagation and enhance relevant feature signals, we process the concatenated features $\tilde{r}_{i-1}$ through a Gated CNN submodule, which is illustrated in right side of Fig.~\ref{fig:architecture}. 

The process starts by expanding the channel of the concatenated feature via a linear projection $Linear$, then splitting it into three parts along the channel dimension:
\begin{equation}
[g,\, i,\, c] = Linear(\tilde{r}_{i-1})
\label{eq:linear_proj}
\end{equation}

Specifically, $Linear$ expands the input from $\mathbb{R}^{h \times w \times d}$ to $\mathbb{R}^{h \times w \times 2\gamma d}$, where the hyperparameter $\gamma$ denotes the expansion ratio (set to $8/3$ in our implementation), and $\alpha$ denotes the convolution ratio (set to $1.0$). The expanded feature with channel dimension $2\gamma d$ is then split along the channel dimension into three parts, respectively:
\begin{itemize}[leftmargin=2em]
    \item g~$\in \mathbb{R}^{h \times w \times \gamma d}$: fed into a \textit{gating path} to generate channel-wise gating weights via GELU activation;
    \item i~$\in \mathbb{R}^{h \times w \times (\gamma-\alpha)d}$: passed through a \textit{global path} to preserve contextual information;
    \item c~$\in \mathbb{R}^{h \times w \times \alpha d}$: processed by a \textit{local path} using standard convolution to extract fine-grained details.
\end{itemize}

The outputs from the global path $i$ and local path $\text{Conv}(c)$ are first concatenated along the channel dimension and then modulated via element-wise multiplication $\odot$ with the gating weights $\text{GELU}(g)$, resulting in the modulated signal $s$:
\begin{equation}
s = \text{GELU}(g) \odot \mathrm{Concat}\left(i,\, \text{Conv}(c)\right)
\label{eq:gating}
\end{equation}

This modulated signal $s$ is linearly projected back to $\mathbb{R}^{h \times w \times d}$ and added to the original input $\tilde{r}_{i-1}$ via a residual connection. A $1\times1$ convolution is then applied to yield the output of current layer:
\begin{equation}
\begin{aligned}
  r_{i-1}^{\text{out}} &= \text{Conv}\left( \tilde{r}_{i-1} + Linear(s) \right)
\end{aligned}
\label{eq:gated_conv}
\end{equation}
The resulting output $ r_{i-1}^{\text{out}} $ is a denoised and semantically enhanced representation, providing a cleaner and more informative signal for the shallower-level refinement in the subsequent iteration, where $r_{i-1}^{\text{out}}$ is fed as $r_{i-1}^{\text{in}}$ at the next stage.
\begin{equation}
r_{i-1}^{\text{in}} \leftarrow r_{i-1}^{\text{out}}
\label{eq:iterative_update}
\end{equation}

This process iterates progressively through all levels $(r_4 \rightarrow r_3 \rightarrow r_2 \rightarrow r_1 \rightarrow r_0 )$. By refining features across multiple levels, the network gradually aggregates and enhances contextual information, ultimately producing a robust, map-relevant satellite map feature representation $r_0^{\text{out}}$. 
To align with the BEV feature map in channel dimensions, this feature is then passed through a convolutional block $\text{Conv}(\cdot)$ that increases the number of channels:
\begin{equation} \label{eq:satellite_bev_output}
F_{\text{sat}} = {\text{Conv}}(r_0^{\text{out}}) \quad \in \mathbb{R}^{C \times H \times W}
\end{equation}


As shown in Fig.~\ref{fig:gate_multi_scale_satellite_map_feature_refinement}, before refinement (mid), strong activations are present in occlusions, shadows, and irrelevant regions outside the road, often caused by vegetation and buildings, indicating noise. After refinement (right), high activations are distributed around the HD map Ground Truth and their surroundings, while noise is largely suppressed. This demonstrates that our gated feature refinement module aligns input with BEV feature and effectively filters out irrelevant signals and enhances map-relevant features, resulting in richer representations for subsequent fusion.

Note that the parameters of the Gated CNN are optimized by minimizing the discrepancy between the input and ground truth. The design of \textit{GELU} activation function enables the network to dynamically adjust the importance of different feature channels by leveraging backpropagation and gradient descent. Further training details are provided in the Evaluation Section.

\begin{figure}[!t] 
\centering
\includegraphics[width=1.0\linewidth]{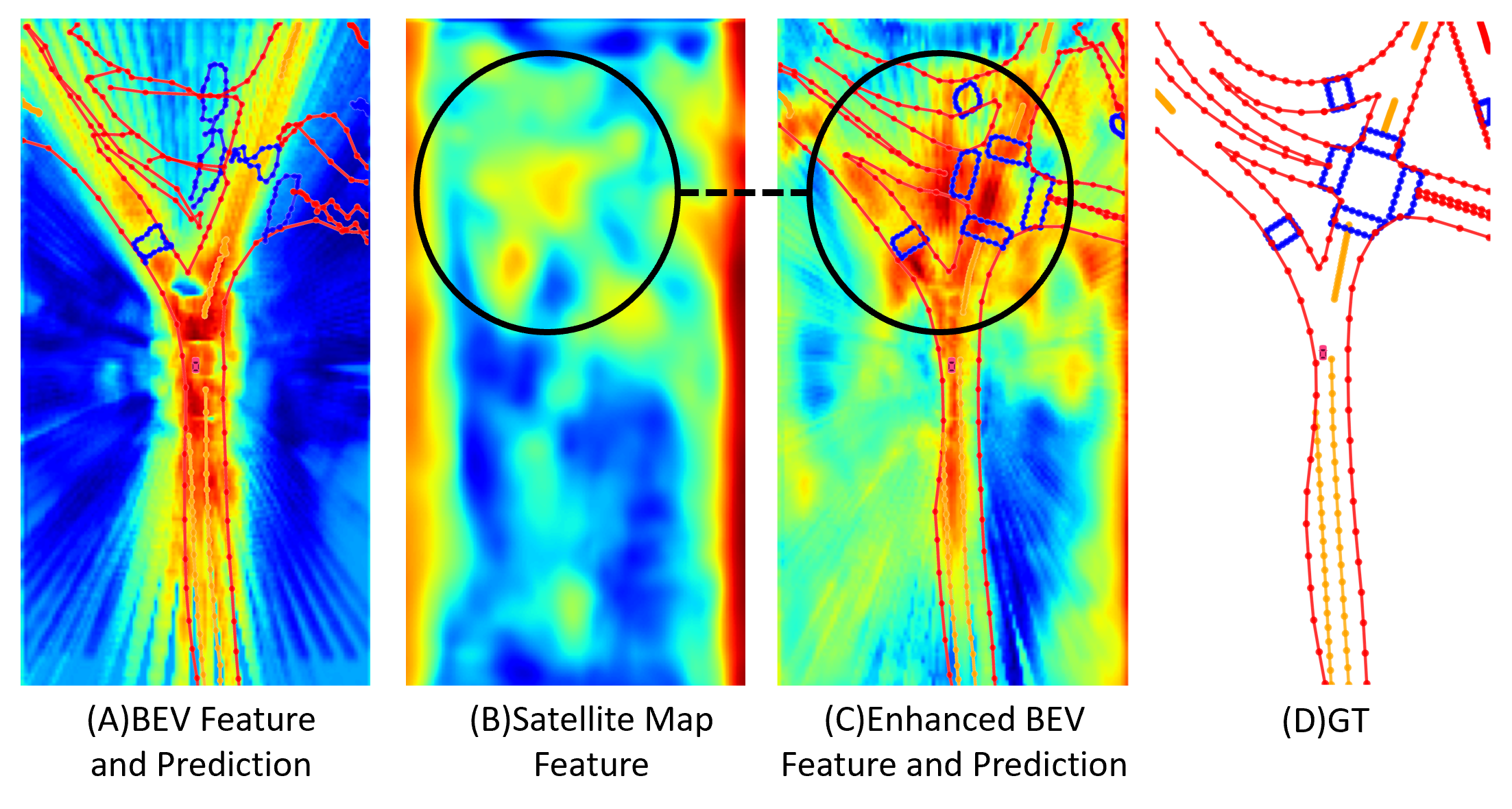}
\caption{Visualization of the Geometry-aware Fusion Module. (A) shows that using only BEV features leads to inaccurate predictions. (B) demonstrates that through the geometry-aware fusion module, high-response regions in the satellite map (circled) are effectively integrated into the enhanced BEV features in (C), resulting in more accurate predictions that align closely with the ground truth (GT) in (D).}
\label{fig:Geometry-aware Fusion Module}
\end{figure}

\subsection{Geometry-Aware Fusion Module}
\label{sec:sat_bev_fusion}
To address Challenge 2, this module consistently fuse the BEV features and the satellite map features with spatial constraints.
Specifically, we propose a geometry-aware fusion mechanism that operates on a per-grid basis to preserve spatial correspondence. For each grid cell, we first perform element-wise summation of the BEV features $F_{\text{bev}}$ and the satellite map features $F_{\text{sat}}$, ensuring that information is fused only at corresponding spatial locations and preventing cross-region information leakage. Notably, this operation is agnostic to input feature quality, ensuring consistent fusion even when inputs are degraded. The resulting fused features are then passed through a Multi-Layer Perceptron (MLP) for further optimization. Importantly, the MLP is applied independently to each grid cell, maintaining spatial isolation and enhancing the map relevance of the fused representation. The entire process is summarized as:
\begin{equation}
\label{eq:enhanced}
F_{\text{enhanced}} = \text{MLP}(F_{\text{bev}} + F_{\text{sat}})
\end{equation}

This design ensures that fusion is both spatially constrained and sufficiently expressive, enabling the learning of more accurate, geometry-aware representations. Finally, the enhanced BEV feature $F_{\text{enhanced}}$ is forwarded to the decoder for HD map prediction. 
As shown in Fig.~\ref{fig:Geometry-aware Fusion Module}, when only the BEV feature (A) is forwarded to the decoder, the resulting HD map predictions are often noisy and less reliable. By using Geometry-aware fusion to fuse the BEV feature with the satellite map feature (B), we obtain the enhanced BEV feature (C). The black circled region illustrates that the enhanced BEV feature correctly integrates BEV and satellite map information within the corresponding area, without interference from irrelevant features. Moreover, As the enhanced BEV Prediction mostly matches ground truth (GT), the regions around the GT(D) map elements exhibit even higher activation in the enhanced BEV feature compared to the original inputs, enabling the decoder to produce more accurate and ground-truth-aligned HD map predictions. This demonstrates that our module can effectively fuse and optimize the representation towards map-relevant and decoder-friendly features. 

\section{Evaluation} 
\subsection{Experiment Setting}
\subsubsection{Datasets and Benchmarks.}
We evaluate SATMapTR on a large-scale real-world autonomous driving dataset \textit{nuScenes}~\cite{caesar2020nuscenesmultimodaldatasetautonomous} and its complementary \textit{nuScenes satellite image dataset}~\cite{gao2024complementingonboardsensorssatellite}, which contain 1,000 scenes, each comprising a 20-second driving sequence. Each frame includes RGB images from six cameras covering 360 degrees horizontal FOV of the ego-vehicle. Following the official nuScenes split, we use 700 scenes for training and 150 scenes for evaluation. In addition, we employ tools provided by MapBench~\cite{hao2024hdmapconstructorreliable} to simulate adverse scenarios, including fog, snow, FrameLost (randomly drop images), CameraCrash (drop images from fixed cameras), and low light conditions.

Following prior works~\cite{li2022hdmapnetonlinehdmap, liao2022maptr, liu2023vectormapnetendtoendvectorizedhd}, we evaluate using Average Precision (AP) over Chamfer Distance thresholds $\mathrm{T} = \{0.5, 1.0, 1.5\}$:
$\mathrm{AP} = \frac{1}{|\mathrm{T}|} \sum_{\tau \in \mathrm{T}} \mathrm{AP}_\tau$. A prediction is a true positive if it matches a ground-truth instance within distance threshold $\tau$. We report AP for lane dividers ($\mathrm{AP}_\mathit{div.}$), pedestrian crossings ($\mathrm{AP}_\mathit{ped.}$), and road boundaries ($\mathrm{AP}_\mathit{bou.}$), and compute mean AP ($\mathrm{mAP}$) as the average across all categories.

\subsubsection{Implementations.}
All training, including SATMapTR, reproduced baselines, and ablation studies, is conducted on 2 $\times$ L20 GPUs using the AdamW optimizer~\cite{loshchilov2019decoupledweightdecayregularization} with a cosine annealing learning rate schedule. We adopt ResNet18~\cite{he2015deepresiduallearningimage} as the satellite image backbone, with a batch size of 4 and an initial learning rate of $2 \times 10^{-4}$ for all models train by ourselves, except for those results cited from the original papers. 

All our reproduction and experiments are trained for 24 epochs, except for HDMapNet and SatForHDMap, which is trained for 30 epochs.
The default perception range is $60\mathrm{m}\times30\mathrm{m}$. For long-range evaluation, we compare performance at $60\mathrm{m}\times30\mathrm{m}$, $120\mathrm{m}\times60\mathrm{m}$, and $240\mathrm{m}\times60\mathrm{m}$. For challenging conditions, we adopt $120\mathrm{m}\times60\mathrm{m}$ to better represent real-world driving and assess robustness for distant elements.

\subsubsection{Baseline Models.}
We compare our SATMapTR with state-of-the-art HD map construction models based on onboard sensors, including HDMapNet~\cite{li2022hdmapnetonlinehdmap}, MapTR~\cite{liao2022maptr}, and MapTRv2~\cite{liao2024maptrv2endtoendframeworkonline}. We also include P-MapNet~\cite{jiang2024pmapnetfarseeingmapgenerator} integrates SD map based on MapTR and SatforHDMap~\cite{gao2024complementingonboardsensorssatellite}, which fuses satellite image with BEV features and provides variants built upon HDMapNet and MapTR. Our framework is implemented on MapTRv2, and also implemented on MapTR for a fair comparison.

\begin{table}
  \centering
  \setlength{\tabcolsep}{1pt} 
  \footnotesize 
  \begin{tabular}{lcccccc}
    \toprule
    Model & Input & $\mathrm{AP}_\mathit{div.}$ & $\mathrm{AP}_\mathit{ped.}$ & $\mathrm{AP}_\mathit{bou.}$ & $\mathrm{mAP}$ & FPS \\
    \midrule
    HDMapNet & C & 29.4 & 26.4 & 45.1 & 33.6 & 0.9 \\
    +SatforHDMap & C+S & 51.6 & 43.0 & 29.2 & 41.2 (+7.6) & 1.6 \\
    \midrule
    MapTR* & C & 51.5 & 46.3 & 53.1 & 50.3 & 15.1 \\
    +SatforHDMap* & C+S & 55.3 & 47.2 & 55.3 & 52.6 (+2.3) & -- \\
    +Ours & C+S & \textbf{67.7} & \textbf{66.2} & \textbf{66.5} & \textbf{66.8 (+16.5)} & -- \\
    \midrule
    MapTR* & C & 49.5 & 41.2 & 51.0 & 47.3 & -- \\
    +P-MapNet* & C+SD & 50.9 & 43.7 & 53.5 & 49.4 (+2.21) & -- \\
    \midrule
    MapTRv2 & C & 61.7 & 58.6 & 62.7 & 61.0 & \textbf{16.3} \\
    MapTRv2* & C & 62.4 & 59.8 & 62.4 & 61.5 & 14.1 \\ 
    MapTRv2* & C+L & 66.5 & 65.6 & 74.8 & 69.0 & 5.8 \\ 
    +Ours & C+S & \textbf{72.3} & \textbf{73.9} & \textbf{75.1} & \textbf{73.8 (+12.8)} & 14.4 \\
    \bottomrule
  \end{tabular}
  \caption{
    Performance comparison of various online HD map construction models under a BEV range of $60\,\mathrm{m} \times 30\,\mathrm{m}$. 
    ``*'' denotes results cited from the original paper, others are our reproduction. 
    C: Camera; S: Satellite; SD: SD map; L: LiDAR.
  }
  \label{tab:performance}
\end{table}

\subsection{Performance Evaluation}
\subsubsection{Accuracy and Computational Efficiency.}


As shown in Table~\ref{tab:performance}, SATMapTR achieves the highest mAP, outperforming MapTRv2 Baseline by 12.8 mAP higher and SatforHDMap by 21.2 higher mAP, highlighting the effectiveness of leveraging satellite imagery to complement pure onboard sensor data. Notably, SATMapTR also exceeds the performance of MapTRv2 with camera and LiDAR inputs. On the same baseline, SATMapTR delivers 14.2 higher mAP over SatforHDMap. Additionally, compared to P-MapNet with SD map, SATMapTR achieves 17.4 mAP higher, demonstrating the advantage of using satellite input. These gains are achieved with real-time inference and only a minor 1.9 FPS decrease, demonstrating that our advanced feature refinement and fusion strategies deliver higher accuracy under standard scenarios while preserving efficiency.

\begin{figure} 
\includegraphics[width=1.0\linewidth]{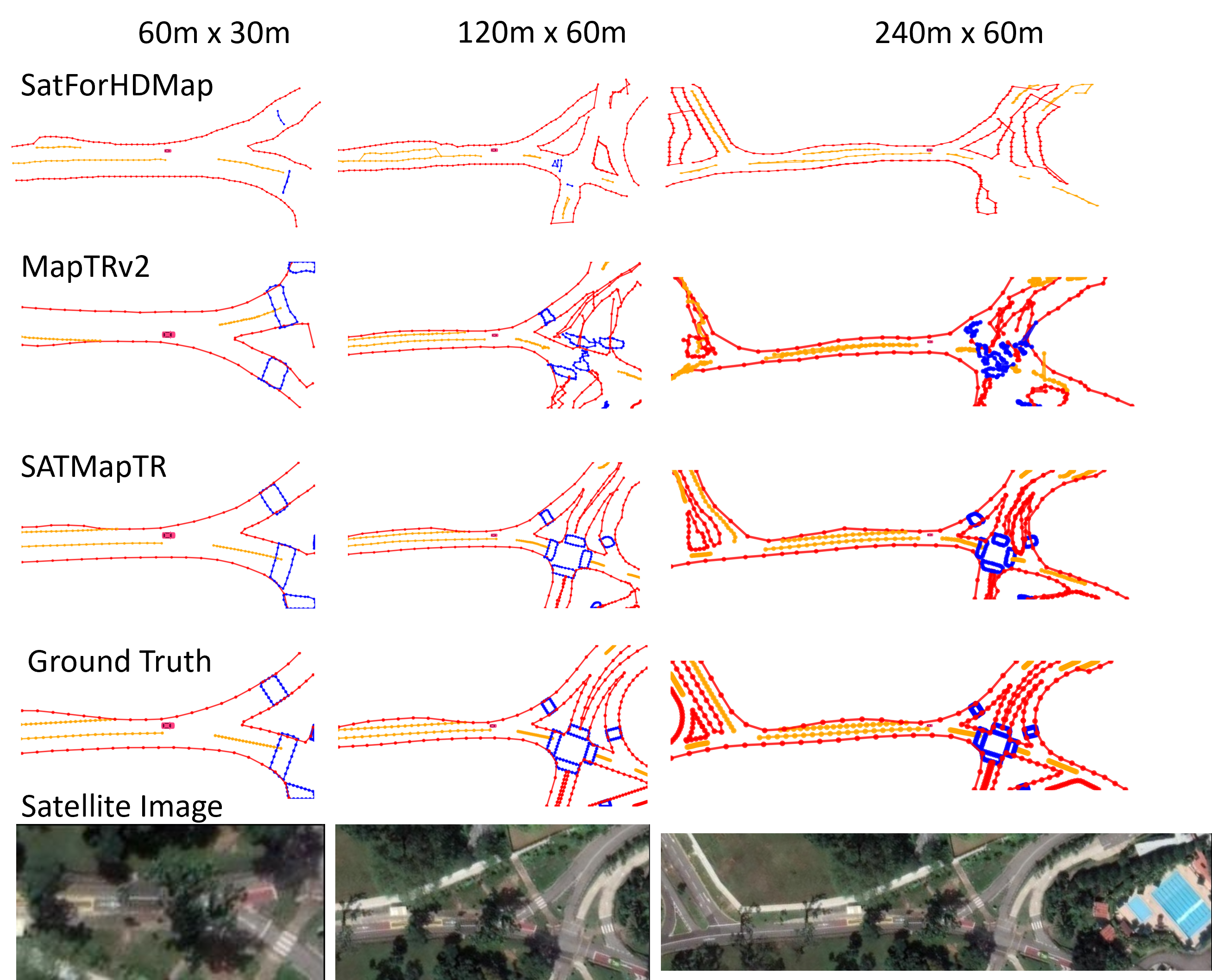}
\centering
\caption{Visualization of HD map Predition under different perception ranges.}
\label{fig:HDMap_Prediction_in_Different_perception_ranges}
\end{figure}

\subsubsection{Performance across Perception Ranges.}
We further evaluate our method and baselines under different perception ranges. As shown in Table~\ref{tab:performance_bev_range}, all models show a relative mAP decrease as the perception range increases, but SATMapTR consistently achieves the highest accuracy with the smallest drop. At the mid range ($120\mathrm{m}\times60\mathrm{m}$), SatforHDMap, MapTRv2, and SATMapTR show relative mAP drops of 62.4

This advantage is more pronounced at the long range ($240\mathrm{m}\times60\mathrm{m}$), where MapTRv2's mAP drops to the same level as SatforHDMap, revealing its limitation in modeling distant features and the satellite image input ensures a lower bound performance. These results indicate that our model effectively integrates BEV and satellite map features to ensure robust performance across all perception ranges. The corresponding visualizations are shown in Fig.~\ref{fig:HDMap_Prediction_in_Different_perception_ranges}.

\begin{table}
  \centering
  \setlength{\tabcolsep}{1pt} 
  \footnotesize
  \begin{tabular}{llcccc}
    \toprule
    \begin{tabular}[c]{@{}l@{}}Range\end{tabular} & Model & $\mathrm{AP}_\mathit{div.}$ & $\mathrm{AP}_\mathit{ped.}$ & $\mathrm{AP}_\mathit{bou.}$ & $\mathrm{mAP}$ \\
    \midrule
    $60 \times 30$ & SatforHDMap    & 51.6 & 43.0 & 29.2 & 41.2 \\
                  & MapTRv2$^*$    & 62.4 & 59.8 & 62.4 & 61.5 \\
                  & Ours           & \textbf{72.3} & \textbf{73.9} & \textbf{75.1} & \textbf{73.8} \\
    \midrule
    $120 \times 60$ & SatforHDMap   & 26.7 & 11.9 & 8.0 & 15.5 (--62.4\%) \\
                   & MapTRv2       & 38.9 & 32.6 & 25.1 & 32.2 (--47.7\%) \\
                   & Ours          & \textbf{55.4} & \textbf{48.4} & \textbf{42.1} & \textbf{48.7} (--34.0\%) \\
    \midrule
    $240 \times 60$ & SatforHDMap   & 13.8 & 11.9 & 7.7 & 11.1 (--73.0\%) \\
                   & MapTRv2       & 16.3 & 12.7 & 5.4 & 11.5 (--81.3\%) \\
                   & Ours          & \textbf{41.0} & \textbf{34.3} & \textbf{24.2} & \textbf{33.2} (--55.0\%) \\
    \bottomrule
  \end{tabular}
  \caption{
     Performance comparison under different perception ranges (in meters, i.e., $60 \times 30$ means $60\,\mathrm{m} \times 30\,\mathrm{m}$). Values in parentheses indicate relative $\mathrm{mAP}$ drop percentage with respect to $60\,\mathrm{m} \times 30\,\mathrm{m}$; “$^*$” denotes results cited from the original paper, others are our reproduction.
  }
  \label{tab:performance_bev_range}
\end{table}

\begin{table}[!t]
  \centering
  \setlength{\tabcolsep}{2.0pt} %
  \footnotesize
  \begin{tabular}{l l c c c c}
    \toprule
    Scenario & Model & $\mathrm{AP}_\mathit{div.}$ & $\mathrm{AP}_\mathit{ped.}$ & $\mathrm{AP}_\mathit{bou.}$ & $\mathrm{mAP}$ \\
    \midrule
    Normal      & MapTRv2 & 38.9 & 32.6 & 25.1 & 32.2 \\
                         & Ours    & \textbf{55.4} & \textbf{48.4} & \textbf{42.1} & \textbf{48.7} \\
    \midrule
    Fog         & MapTRv2 & 36.1 & 29.6 & 22.3 & 29.4 (–2.8) \\
                         & Ours    & \textbf{55.0} & \textbf{47.2} & \textbf{41.2} & \textbf{47.8} (–0.9) \\
    \midrule
    Snow        & MapTRv2 & 2.7  & 0.8  & 0.6  & 1.4 (–30.8) \\
                         & Ours    & \textbf{41.6} & \textbf{34.5} & \textbf{25.5} & \textbf{33.9} (–14.8) \\
    \midrule
    FrameLost   & MapTRv2 & 29.2 & 24.5 & 16.8 & 23.5 (–8.7) \\
                         & Ours    & \textbf{51.5} & \textbf{44.1} & \textbf{36.6} & \textbf{44.1} (–4.6) \\
    \midrule
    CameraCrash & MapTRv2 & 35.9 & 30.0 & 22.0 & 29.3 (–2.9) \\
                         & Ours    & \textbf{54.7} & \textbf{46.9} & \textbf{40.7} & \textbf{47.4} (–1.3) \\
    \midrule
    LowLight    & MapTRv2 & 18.3 & 14.2 & 8.9  & 13.8 (–18.4) \\
                         & Ours    & \textbf{52.0} & \textbf{44.5} & \textbf{36.3} & \textbf{44.2} (–4.4) \\
    \bottomrule
  \end{tabular}
  \caption{
    Performance comparison in challenging conditions under a BEV range of $120\,m \times 60\,m$. Values in parentheses indicate $\mathrm{mAP}$ drop compared to the normal scenario. 
  }
  \label{tab:performance_challenging_condition}
\end{table}


\begin{table}
  \centering
  \setlength{\tabcolsep}{6pt}
  \footnotesize
  \begin{tabular}{lcccc}
    \toprule
    \diagbox{$\sigma_r$ (rad)}{$\sigma_t$ (m)} & 0 & 0.05 & 0.1 & 0.2 \\
    \midrule
    0     & 73.75 & 73.16 & 73.02 & 72.02 \\
    0.005 & 72.96 & 72.92 & 72.83 & 71.88 \\
    0.01  & 72.68 & 72.53 & 72.38 & 71.43 \\
    0.02  & 70.66 & 70.64 & 70.51 & 69.61 \\
    \bottomrule
  \end{tabular}
  \caption{  
    Performance with different localization errors. Gaussian noise is added to translation ($\sigma_t$) and rotation ($\sigma_r$) components of the ego-pose as localization errors and test with same model. 
  }
  \label{tab:loc_error_map}
\end{table}

\subsubsection{Performance Under Challenging Conditions.}
As shown in Table~\ref{tab:performance_challenging_condition}, SATMapTR not only maintains much smaller mAP drops compared to normal conditions, but also achieves significantly higher mAP than MapTRv2 across all challenging scenarios—even surpassing the normal-scenario performance of MapTRv2. In Snow, FrameLost, and CameraCrash, MapTRv2 suffers mAP declines more than twice those of SATMapTR, while in Fog and Low-light, its mAP drops are approximately three and four times greater, respectively. 

These results clearly demonstrate SATMapTR’s robustness and reliability under adverse weather, sensor failures, and low-light conditions. (see supplementary material~\ref{sec:challenging_analysis} for visualizations and detailed analysis.)

\subsubsection{Robustness to Localization Error}
As described in the Method section, satellite image alignment relies on the ego-vehicle's pose for coarse alignment via GNSS/INS, and compensation by convolution blocks in GFR. However, real-world errors may arise from environmental factors. We assess SATMapTR's robustness on nuScenes (Table ~\ref{tab:loc_error_map}) by adding Gaussian noise to translation ($\sigma_t$) and rotation ($\sigma_r$) in ego-pose.

Results indicate strong robustness, especially to translation noise. For $\sigma_r = 0$, mAP drops modestly from 73.75 ($\sigma_t = 0$) to 72.02 ($\sigma_t = 0.2$ m). At maximum noise ($\sigma_t = 0.2$ m, $\sigma_r = 0.02$ rad), mAP is 69.61, still outperforming MapTRv2 baseline (61.5 mAP). This stems from the convolution blocks in GFR module, which compensates for misalignments (see Fig.~\ref{fig:gate_multi_scale_satellite_map_feature_refinement}); In most cases, the mAP drops are under 2$\%$ relative to the noise-free baseline.

Given that 0.1 m translation and 0.01 rad rotation errors are typical requirements in autonomous driving \cite{Reid2019LocalizationRF}, SATMapTR ensures reliable HD map construction in practice.

\subsection{Ablation Study Results}
\subsubsection{Gated Feature Refinement Module}
As shown in Table~\ref{tab:ablation}, by using the same fusion methods, directly fusing satellite features without refinement brings only a slight mAP improvement of 1.1. In contrast, introducing the complete GFR module boosts mAP by 12.3, highlighting the importance of multi-level gated feature refinement. Using a standard convolution to replace the Gated CNN (GFR$^*$) yields a 10.0 mAP gain, which is notably less than the full GFR. These results demonstrate that both geometry-aware fusion and the full GFR module are crucial for optimal accuracy, with GFR being especially important for extracting high-quality map features.

\begin{table}
  \centering
  \setlength{\tabcolsep}{2pt}
  \footnotesize
  \begin{tabular}{lcc|cccc}
    \toprule
    Model & GFR$^*$ & GFR & $\mathrm{AP}_\mathit{div.}$ & $\mathrm{AP}_\mathit{ped.}$ & $\mathrm{AP}_\mathit{bou.}$ & $\mathrm{mAP}$ \\
    \midrule
    MapTRv2$^*$ & $-$ & $-$ & 62.4 & 59.8 & 62.4 & 61.5 \\
    \midrule
                & $\times$ & $\times$ & 63.4 & 61.1 & 63.3 & 62.6 (+1.1) \\
                & $\checkmark$ & $\times$ & 70.3 & 71.7 & 72.6 & 71.5 (+10.0) \\
    Ours        & $\times$ & $\checkmark$ & \textbf{72.3} & \textbf{73.9} & \textbf{75.1} & \textbf{73.8} (\textbf{+12.3}) \\
    \bottomrule
  \end{tabular}
  \caption{
    Ablation results for the Gated Feature Refinement (GFR) module. (``GFR$^*$'' denotes the GFR module without the Gated CNN block; ``GFR'' denotes the complete GFR module. “$^*$” denotes results cited from the original paper.)
  }
  \label{tab:ablation}
\end{table}


\begin{table}
  \centering
  \setlength{\tabcolsep}{1pt} 
  \footnotesize
  \begin{tabular}{lccccc}
    \toprule
    Model & Fusion & $\mathrm{AP}_\mathit{div.}$ & $\mathrm{AP}_\mathit{ped.}$ & $\mathrm{AP}_\mathit{bou.}$ & $\mathrm{mAP}$ \\
    \midrule
    MapTR$^*$ & -- & 51.5 & 46.3 & 53.1 & 50.3 \\
    SatforHDMap$^*$ & Attn. & 55.3 & 47.2 & 55.3 & 52.6 (+2.3) \\
    Ours & PatchCA & 39.0 & 28.4 & 38.4 & 35.3 (-15.0) \\
    Ours & Sum & \textbf{67.7} & \textbf{66.2} & \textbf{66.5} & \textbf{66.8} (+\textbf{16.5}) \\
    \midrule
    MapTRv2$^*$ & -- & 62.4 & 59.8 & 62.4 & 61.5 \\
    Ours & PatchCA & 46.9 & 37.2 & 46.5 & 43.5 (–18.0) \\
    Ours & Concat & 69.0 & 72.0 & 73.4 & 71.5 (+10.0) \\
    Ours & Sum & \textbf{70.3} & \textbf{71.7} & \textbf{72.6} & \textbf{71.5} (+\textbf{10.0}) \\
    \bottomrule
  \end{tabular}
  \caption{
    Comparison of different fusion methods. All models are based on the group header. "$^*$" denotes results cited from the original paper.
  }
  \label{tab:fusion_comparison}
\end{table}

\subsubsection{Effectiveness of Different Fusion Methods}
We evaluate various fusion strategies using the same satellite image feature extractor and baseline models within our framework to ensure a fair comparison. For attention-based methods, we follow the ViT paradigm to divide input feature into patch embeddings to reduce computation. As shown in Table~\ref{tab:fusion_comparison}, we use SatforHDMap (based on MapTR) as a reference for attention-based fusion with soft constraints (Attn.), which yields only a marginal improvement of 2.3 mAP over MapTR. Patch-based cross-attention without constraint (PatchCA) even results in a significant 15.0 mAP drop, indicating that unconstrained global attention can introduce irrelevant dependencies and disrupt local feature integrity.

In contrast, simple element-wise summation or concatenation achieves the highest gains, with the sum method improving mAP by 16.5 on MapTR and 10.0 on MapTRv2.

Moreover, As shown in Fig.~\ref{fig:satforhdmap_challenging_comparison}, SatforHDMap partially recovers HD maps at intersection under normal conditions, and its performance further drops with low-quality onboard input in fog, where satellite features provide little compensation. This exposes a key limitation of attention-based fusion: when onboard inputs degrade, BEV features lack essential map elements, thus it cannot effectively match corresponding map elements and aggregate them from complementary features. In contrast, SATMapTR consistently produces near-ground-truth predictions even under challenging conditions, demonstrating that explicit, localized feature interactions are not only simpler but also more robust and effective for maintaining feature fidelity and overall performance.

\section{Conclusion}
This paper presents SATMapTR, a novel end-to-end satellite-BEV fusion framework for online vectorized HD map construction. 
It incorporates a satellite map feature extraction module that effectively suppresses noise in satellite imagery and a geometry-aware fusion module that performs reliable grid-to-grid fusion under spatial constraints. Extensive experiments on the nuScenes dataset demonstrate that the proposed method outperforms state-of-the-art vision-based and satellite-enhanced solutions. It also achieves superior performance at larger perception ranges and maintains robustness under challenging conditions, such as adverse weather and sensor failures.

\clearpage  

{
    \small
    \bibliographystyle{ieeenat_fullname}
    \bibliography{main}
}

\clearpage
\setcounter{page}{1}
\maketitlesupplementary


\begin{figure}
\centering
\includegraphics[width=1.0\linewidth]{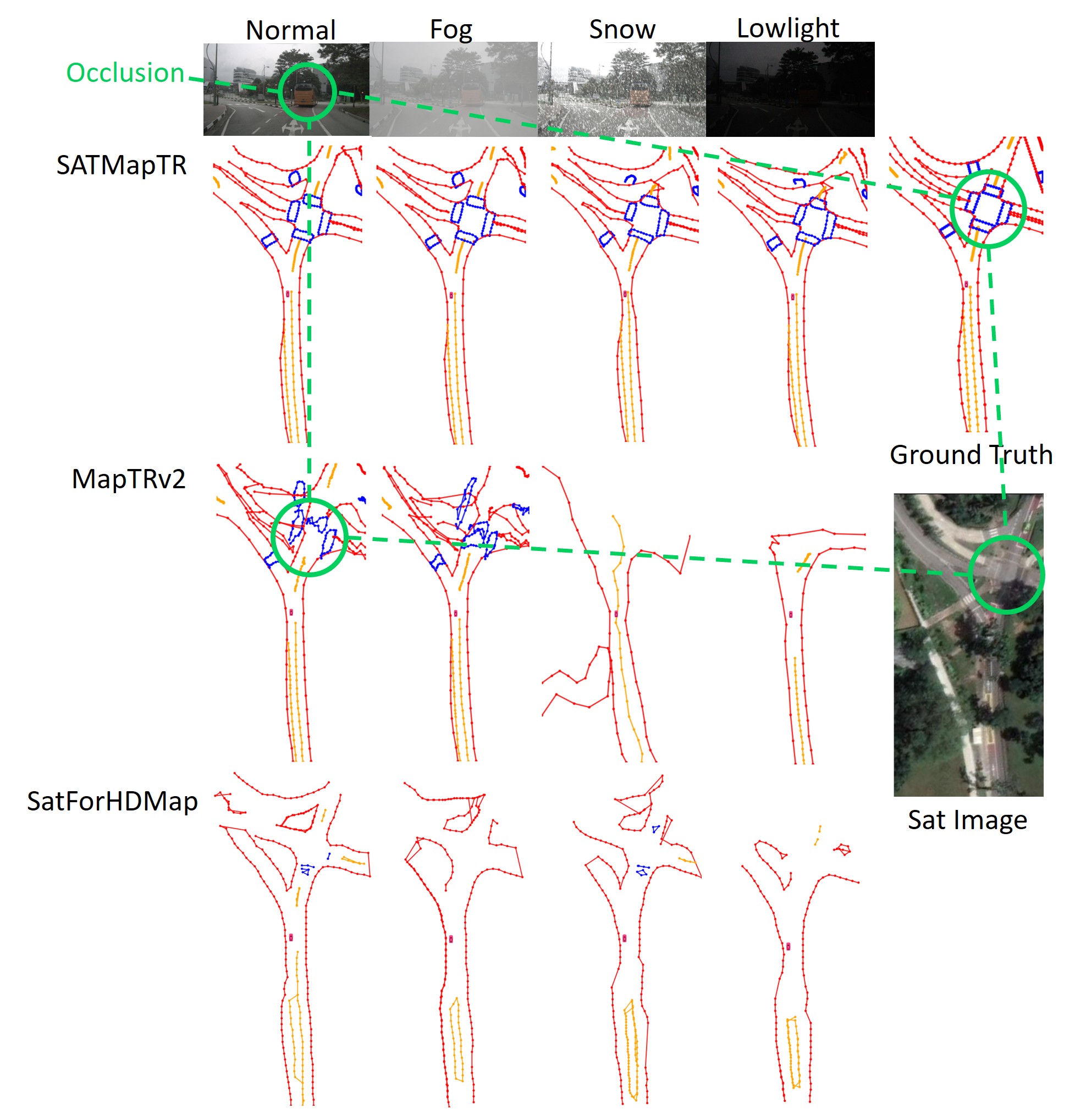}
\caption{Comparison of HD map construction results of different models under normal and different challenging conditions in perception range $120\,\mathrm{m} \times 60\,\mathrm{m}$. Green circle region indicates the occlusion and corresponding area in HD map and satellite image.}
\label{fig:challenging_comparison}
\end{figure}
\section{HD Map Predictions in Challenging Conditions}
\label{sec:challenging_analysis}

we present HD map prediction results of different methods in a challenging intersection scenario with occlusion from a preceding vehicle, under both normal and adverse conditions (such as fog, snow, and low light).

Our method employs a sum-based fusion under precise geometric alignment: BEV features and satellite features at the same physical location are directly added and jointly optimized towards HD map elements. This simple but tightly constrained operation ensures that the fused representation at each grid cell consistently encodes information from both modalities at the corresponding spatial position. As a result, the model can retain complementary information even when one modality is degraded (e.g., noisy BEV features in low-light conditions or partially occluded regions). Empirically, this yields more complete and accurate HD map predictions across a variety of adverse scenarios, demonstrating the robustness and effectiveness of our fusion strategy.

In contrast, purely camera-based MapTRv2 is highly sensitive to occlusions and adverse environments. Due to occlusion from the preceding vehicle on the right side of the road (green circle), MapTRv2 can only predict the left-side crosswalk under normal conditions and essentially fails to recognize map elements at the intersection. This issue becomes more severe under challenging conditions: in snow and low-light scenarios, MapTRv2 can hardly predict any HD map elements within the perception range.

SatforHDMap improves over MapTRv2 by incorporating satellite imagery, which enables basic recognition of coarse map geometry under normal conditions, such as road contours and road boundaries. However, it still struggles to accurately identify fine-grained road elements, including lane dividers and pedestrian crossings, and therefore fails to generate predictions that closely match the ground truth. This performance gap becomes more pronounced as the prediction range increases, where distant map elements are already difficult to perceive from onboard sensors alone. Under challenging conditions, especially in low-light scenarios or in the presence of occlusions, SatforHDMap degrades further: it frequently misses distant or partially occluded map elements and often produces fragmented or inconsistent structures, indicating that its attention-based BEV–satellite fusion does not fully exploit the complementary information provided by the satellite modality.

\begin{figure}[t]
\centering
\includegraphics[width=1.0\linewidth]{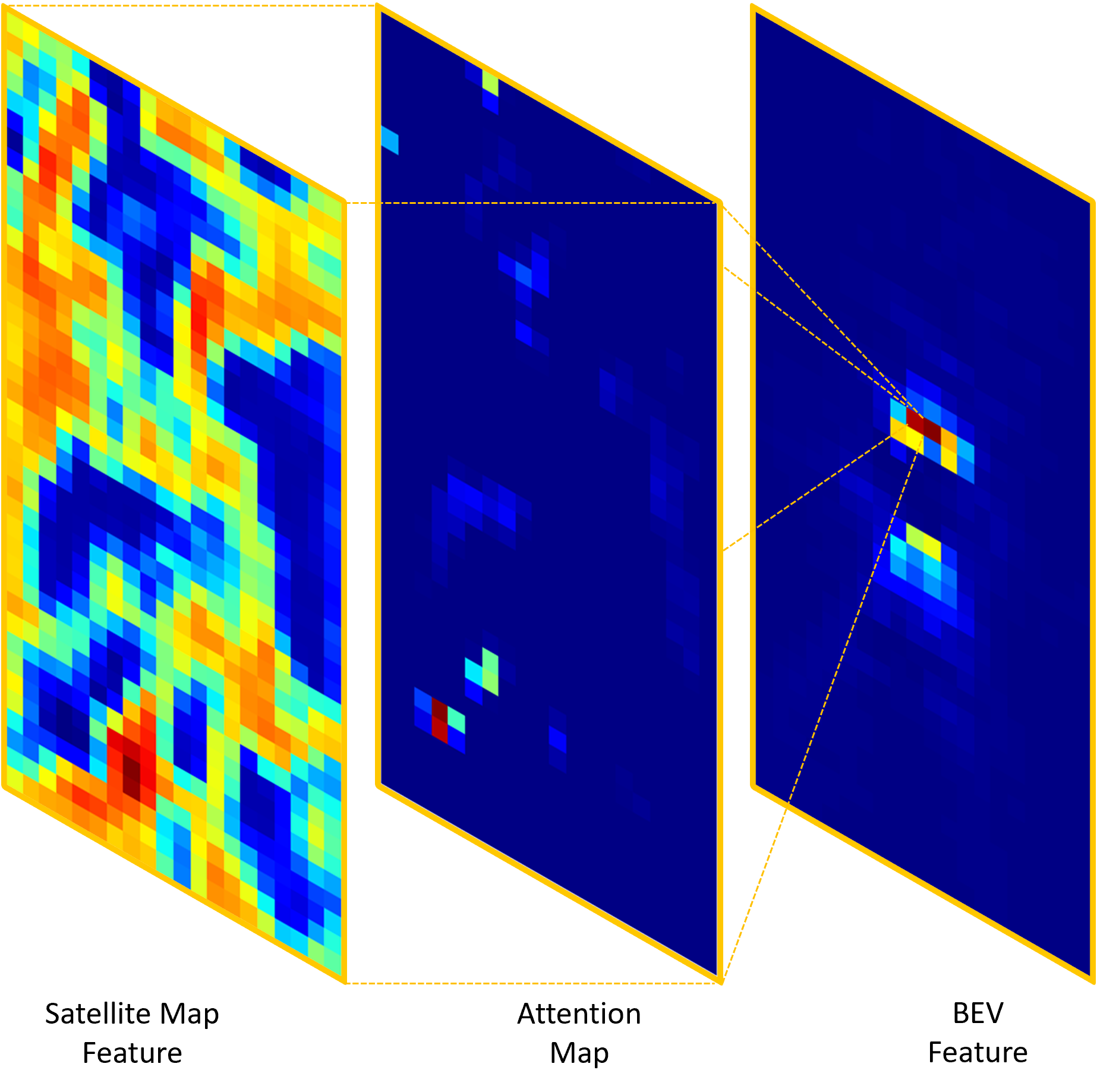}
\caption{
Illustration of cross-attention between a BEV feature grid cell (right) and satellite map features (left), where the attention map (middle) shows the attention weights from the selected BEV feature vector to the satellite map features.
}
\label{fig:cross_atten}
\end{figure}

\section{Limitations of Global Cross-Attention in Early Fusion}
\label{sec:limitation_ca}
The above observations show that attention-based BEV–satellite fusion does not reliably provide complementary benefits in challenging conditions. We next analyze why this happens and why our fusion strategy is more robust.

A key limitation of attention-based BEV–satellite fusion is its dependence on the quality of BEV features. When occlusions or adverse environments degrade BEV features—especially in distant regions—the fusion process is severely weakened. Even if satellite images in these areas contain rich map information, the lack of corresponding map structures in the BEV representation prevents cross-attention from establishing effective dependencies.

Attention mechanisms fundamentally rely on similarity between queries and keys to decide how information is fused. When one input is of low quality or lacks relevant features (e.g., missing lane dividers or pedestrian crossings in the BEV feature), its projected embeddings (queries and/or keys) become noisy and ambiguous. The attention module is then forced to match these unreliable queries to “most similar” regions in the other feature map, which are often irrelevant or noisy. Instead of focusing on complementary satellite regions, the attention weights may concentrate on background or unrelated areas. As a result, the fused features at those positions fail to capture useful information from either modality, leading to ineffective or even harmful fusion and, ultimately, inferior HD map predictions.

Moreover, conventional cross-attention performs global dependency modeling. As illustrated in Fig.~\ref{fig:cross_atten}, a single BEV grid cell attends to the entire satellite feature map and may aggregate features from distant, non-corresponding locations. This global, unconstrained interaction is problematic in early fusion: fusion is performed before the HD map decoder, and each grid cell in the fused feature map is expected to represent its own physical location in the BEV plane. Global attention can break this assumption by mixing features from unrelated regions, distorting the spatial semantics of the BEV representation.

In early fusion, it is crucial that each grid cell encodes information only about its corresponding physical location. This preserves the BEV definition and ensures compatibility with HD map decoders that operate on spatially well-defined grid cells. Our method enforces this property with a strictly location-wise sum under precise geometric alignment: BEV and satellite features are fused only at matched grid cells, without global mixing. This anchors fusion to local, physically meaningful correspondences, allowing the satellite modality to reliably complement degraded BEV features while avoiding spurious dependencies. As a result, our approach achieves more stable early fusion and yields superior HD map predictions across both normal and challenging conditions.

\begin{figure}[h]
\centering
\includegraphics[width=1.0\linewidth]{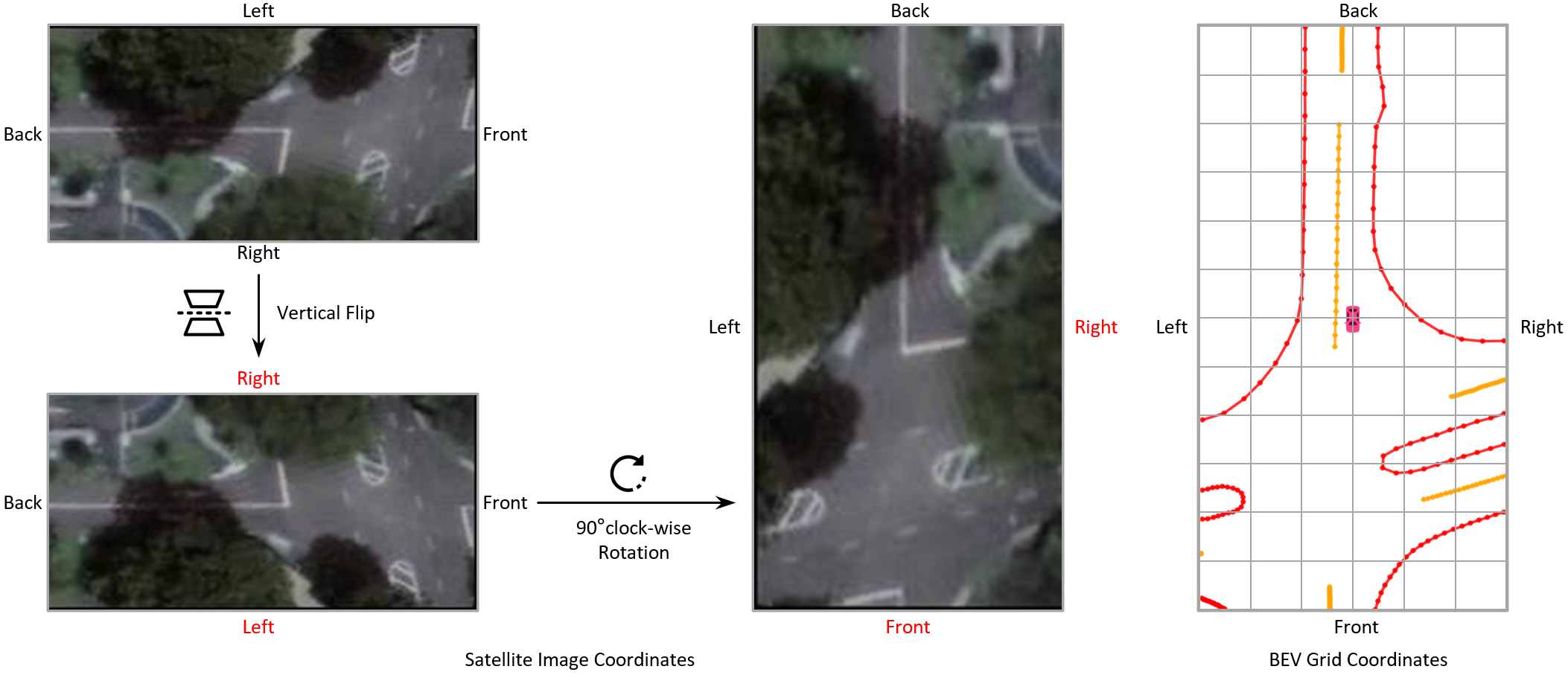}
\caption{
Illustration of aligning the coordinates of an ego-centered satellite image to the BEV coordinate system.
}
\label{fig:convert_satellite_coordinates_to_BEV}
\end{figure}

\section{Satellite Image Alignment}
\label{sec:satalignment}
In this section, we detail how to precisely align satellite map features with BEV features. We first convert the ego-vehicle’s local pose from \textit{nuScenes} into global geographic coordinates (latitude and longitude) and orientation (heading angle). Based on this information, we crop a satellite image patch from a large satellite map, centering it at the ego-vehicle and covering the pre-defined BEV perception range (i.e., $60\,\mathrm{m} \times 30\,\mathrm{m}$ in physical dimensions). The cropped patch is then resized to a fixed resolution ($I_{\text{sat}} \in \mathbb{R}^{H_{\text{sat}} \times W_{\text{sat}} \times 3}$).

Next, we transform the satellite image patch from the global geographic coordinate system to the ego-vehicle coordinate system using an affine transformation. As illustrated in Fig.~\ref{fig:convert_satellite_coordinates_to_BEV}, this transformation consists of a vertical flip followed by a $90^\circ$ clockwise rotation, ensuring that the satellite image and BEV features share the same coordinate system. This process completes the coarse alignment.

To further refine the alignment, we employ a convolutional block within the Gated Feature Refinement module, which aggregates local neighborhood information. Since the features are already coarsely aligned, this local aggregation enables the network to perform fine-grained alignment. The convolution thus serves as an implicit fine alignment, allowing the network to adaptively compensate for small misalignments between the satellite and BEV grids.

\begin{figure}[t]
\centering
\includegraphics[width=1.0\linewidth]{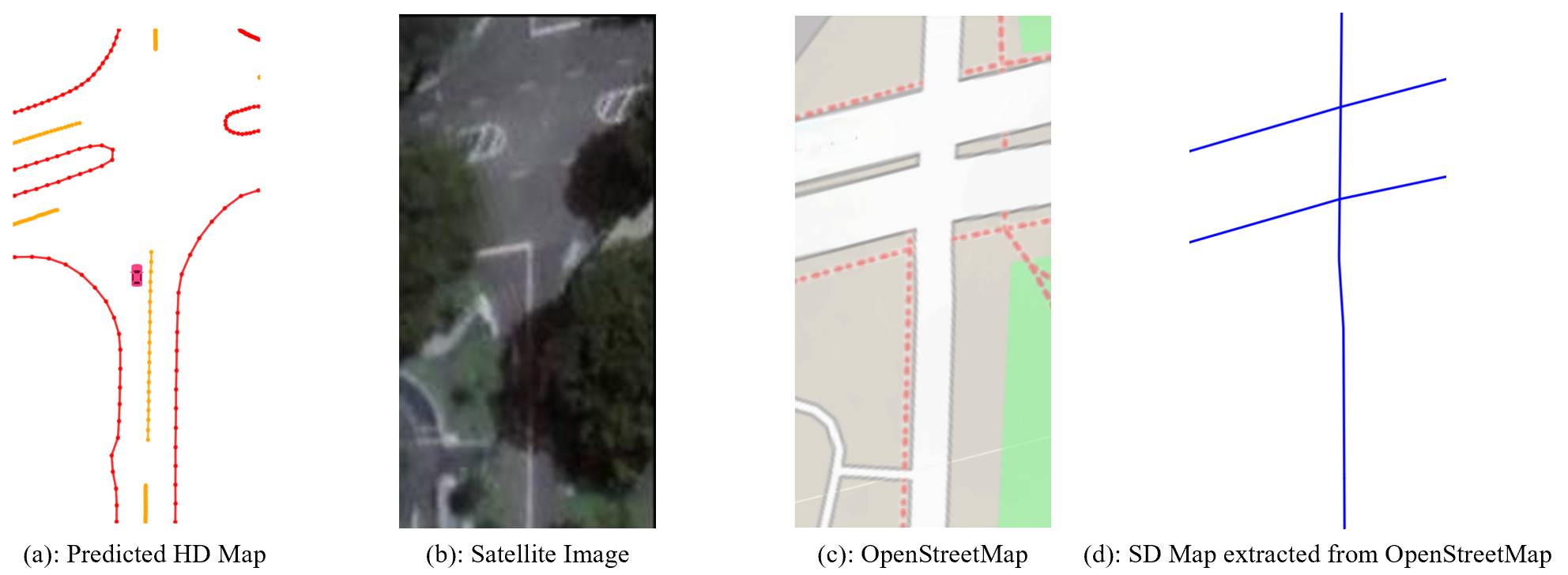}
\caption{
Comparison of the HD map, satellite image, SD map, and raw SD map data.
}
\label{fig:hdmap_sat_sdmap_comparison}
\end{figure}

\section{SD Map and Satellite Image Comparison}
\label{sec:sdsatcom}
As shown in Fig.~\ref{fig:hdmap_sat_sdmap_comparison}, (a) depicts the HD map prediction, which is the target of our task. The satellite image in (b) contains all the corresponding elements required for HD map prediction, including road contours, more precise road boundaries, and lane dividers, all of which can be directly matched to the HD map. This demonstrates that satellite images are rich in the road element information needed for our task.

In contrast, the SD map (c) and its raw data (d) show that the SD map is primarily obtained by further processing the SD map raw data. However, the SD map only provides the general road contour and overall layout, lacking accurate road widths, lane dividers, and crosswalk information. Therefore, the satellite image generally contains as much, if not more, road element information as the SD map.

It should be noted, however, that satellite images itself may suffer from occlusions caused by vegetation, shadows, or buildings. In contrast, the road contours provided by the SD map are more robust and stable under such challenging conditions. In extreme cases, the SD map may still offer reliable guidance on road topology when the satellite image is heavily occluded. Nevertheless, in most scenarios, satellite maps can provide comprehensive coverage of road element information.

\end{document}